# Automatic classification of trees using a UAV onboard camera and deep learning

Masanori Onishi[1]*, Takeshi Ise[2,3]

[1] Graduate School of Agriculture, Kyoto University, Kyoto, Japan
[2] Field Science Education and Research Center, Kyoto University, Kyoto, Japan
[3] PRESTO, Japan Science and Technology Agency, Kawaguchi, Japan.

* corresponding author: onishi.masanori.25e@st.kyoto-u.ac.jp

**Abstract**

Automatic classification of trees using remotely sensed data has been a dream of many scientists and land use managers. Recently, Unmanned aerial vehicles (UAV) has been expected to be an easy-to-use, cost-effective tool for remote sensing of forests, and deep learning has attracted attention for its ability concerning machine vision. In this study, using a commercially available UAV and a publicly available package for deep learning, we constructed a machine vision system for the automatic classification of trees. In our method, we segmented a UAV photography image of forest into individual tree crowns and carried out object-based deep learning. As a result, the system was able to classify 7 tree types at 89.0% accuracy. This performance is notable because we only used basic RGB images from a standard UAV. In contrast, most of previous studies used expensive hardware such as multispectral imagers to improve the performance. This result means that our method has the potential to classify individual trees in a cost-effective manner. This can be a usable tool for many forest researchers and managements.

## 1. Introduction

Automatic classification of individual trees using remotely sensed data has been a dream of many scientists and land use managers. In the recent review, Fassnacht et al. (2016) identified that targets for tree species classification using remote sensing include biodiversity assessment, monitoring of invasive species, wildlife habitat mapping, and sustainable forest management. Various research activities have been carried out to identify tree species, using specialized hardware such as airborne hyperspectral, multispectral, and LiDAR sensors. However, these approaches concerning spectra have often experienced trouble to classify similar spectral species, and also can be affected by shadows and background noises. Therefore, despite the abundance of information contained in hyperspectral imaging, a much restricted set of species can be accurately identified (Peerbhay et al. 2013, Heinzel and Koch 2012).

In recent decades, UAVs have come to be used experimentally for forestry (Tang et al. 2015, Paneque-Gálvez et al. 2014). Comparing to manned aircraft, UAVs are an easy-to-use, cost-effective tool for remote sensing of forests. Moreover, UAVs can fly near canopies and can take higher resolution images; although the images from airplane have spatial resolution of from tens of centimeters to several meters in general, that from UAVs can be a few centimeters.

Meanwhile, deep learning has become a very effective tool for object identification and has shown its high classification performance for digital images. (Krizhevsky et al. 2012, Szegedy et al. 2015, Ise et al. 2017). One of the most notable points is that deep learning do not need manual feature extraction. Previous machine learnings such as support vector machine (SVM) or Random Forest required researchers to choose features or band waves. This process limited the information they can use. In this point, deep learning can use full feature information, so even if we use digital images, deep learning is expected to show a high performance for classification.

From the above, the combination of UAVs photography and deep learning is expected to have a high potential for classifying trees even if we use consumer-grade digital camera. And also, this machine vision system will be a cost-effective and usable tool for forest managements. The objective of this study is to test whether this system can classify individual trees into tree types (deciduous and coniferous) and identify a few specific tree species.

## 2. Material and Methods

*2.1. Study site*

The study site was Kamigamo Experimental Station of Kyoto University, located in a suburban area of Kyoto, Japan. This area belongs to warm and humid climate zone with an elevation of 109 ~ 225 m above sea level. Mean annual precipitation and temperature are 1,582 mm and 14.6 °C, respectively. The overall area is 46.8ha. 65% of the area is natural forest, which consists of mainly Japanese cypress (*Chamaecyparis obtuse)*, and some broad-leaved trees such as oak (*Quercus serrata* or *Quercus glauca).* 28% of the area is planted forest, mainly consisted of foreign coniferous species. 7% is sample garden, nursery or buildings.

In this study, we focused on the northern part (11ha) of the Kamigamo Experimental Station, containing natural forest of Japanese cypress, and managed forest of Metasequoia (*Metasequoia glyptostroboides*), strobe pine (*Pinus strobus)*, slash pine (*Pinus elliottii)* and taeda pine *(Pinus taeda).*

*2.2. Remote Sensing Data*

A flight campaign was conducted in 20 November 2016 during fall leaf offset season. We used commercially available UAV DJI Phantom 4. The UAV has an onboard camera, with a 1/2.3 CMOS sensor that can capture red–green–blue (RGB) spectral information. The UAV was operated automatically using DroneDeploy v2.66 application (www.dronedeploy.com, Infatics Inc., San Francisco, United States). Flight parameters were following that both forward and side overlap are 80%, and flight height is 80m from the takeoff ground level. We used 10 ground-control points (GCPs) for enhancing accuracy. From the images taken by UAV, we made an orthomosaic photo and a digital elevation model (DEM) using Agisoft Photoscan Professional v1.3.4 software (www.agisoft.com, Agisoft LLC, St. Petersburg, Russia). An orthomosaic photo is an image that is composed of multiple overhead images and is corrected for perspective and scale. Resolutions of an orthomosaic photo and a DEM were about 5cm and 9cm, respectively.

### 2.3. *Segmentation and Preparing Training Data*

Each tree image extraction method we used is summarized in Figure 1. First, we segmented each tree crown using UAV photography (orthomosaic photo), digital elevation model (DEM) and slope model. Second, we made ground truth map visually. Third, we extracted each tree image with ground truth label. Finally, we used these images to supervised deep learning. The details are as follows.

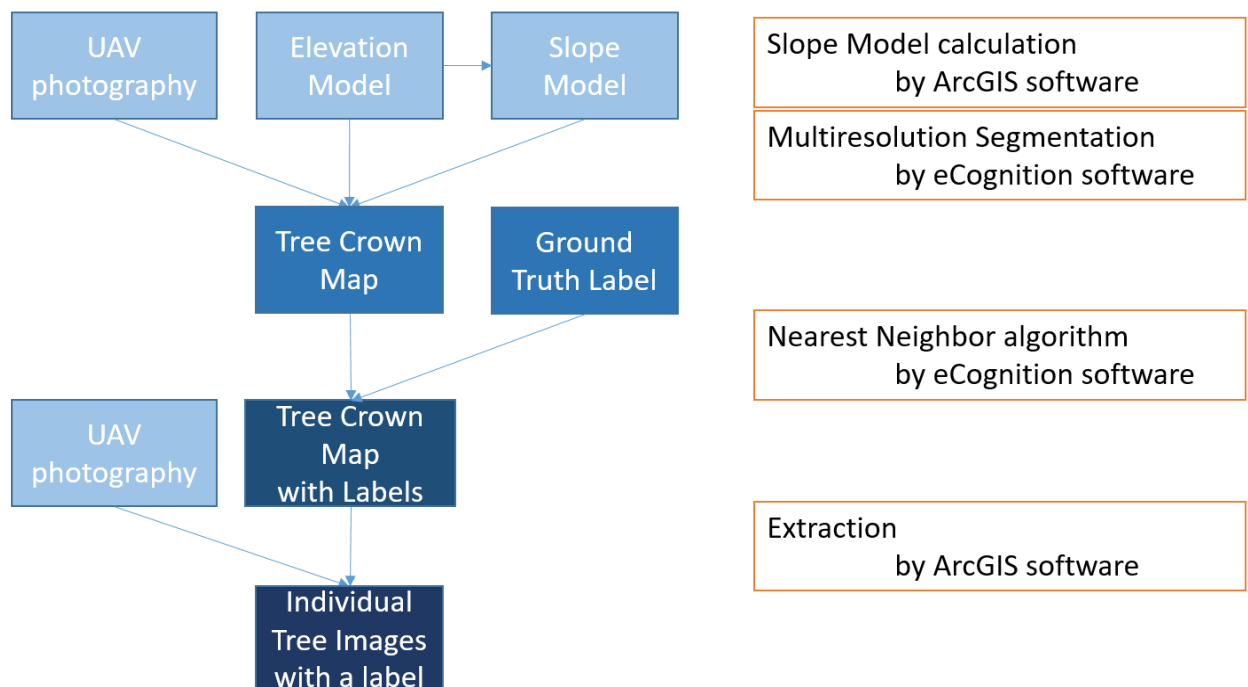

**Figure 1.** Workflow for the training images extraction

### 2.3.1. *Object-Based Tree Crown Segmentation*

From DEM, we made slope model using ArcGIS v10.4 software (Environmental Systems Research Institute, Inc., Redlands, United States). Slope model showed the maximum rate of elevation change between each cell and its neighbors, so the borders of trees were emphasized. From orthomosaic photo, DEM, and slope model, tree crown segmentation was performed in eCognition Developer v9.0.0 software (www.trimble.com, Trimble, Inc., Sunnyvale, United States) using the "Multiresolution Segmentation" algorithm (Baatz & Schäpe 2000). Parameter values were adjusted by try and error. In the best parameter values (Table 1), some segmented images had two or three tree crowns, but this method almost succeeded to separate each class tree. After that, we manually revised some polygons

Table 1. Parameters for multiresolution

| Setting | Selected option |
| --- | --- |
| Weight of R, G, B, DEM, Slope Model | 1, 1, 1, 2, 3 |
| scale | 200 |
| compactness | 0.5 |
| shape | 0.2 |

*2.3.2. Ground Truth Label Attachment to Tree Crown Map*

After segmentation, we classified segmented images into 7 classes: "deciduous broad-leaved tree", "deciduous coniferous tree", "evergreen broad-leaved tree", "*Chamaecyparis obtuse*", "Pinus", "*Pinus strobus*" and "others". Pinus class consisted of *Pinus elliottii* and *Pinus taeda.* The "others" class included understory vegetation between tree crowns and bare land, as well as artificial structures. In eCognition software, using the nearest neighbor classification which is used for forest mapping (Machala & Zejdová 2014), efficiently we made ground truth map (Figure 2). The detail is below that we chose some image objects as training samples visually and applied that algorithm to the overall tree crown map. In subsequent steps, by adding wrongly classified objects to correct classes of training samples, we improved ground truth map accuracy.

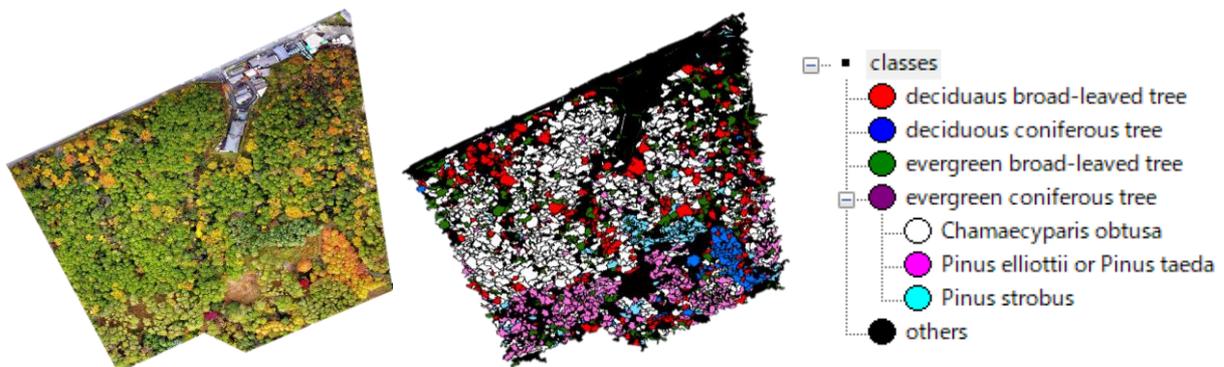

**Figure 2.** Segmentation and Ground Truth Map making result

*2.3.3. Each Tree Image Extraction with Ground Truth Label*

From orthomosaic photo and tree crown map with ground truth label, we extracted each tree image with a label using the "Extract by Mask" function in ArcGIS. There were some inappropriate images such as fragments of trees and including multiple class trees. We manually deleted inappropriate images, and replaced wrongly classified images into a correct class. We show representative images at Figure 3. In the figure, "*Chamaecyparis obtuse*" image has 3 tree crowns and "*Pinus strobus*" was not segmented well. The number of extracted images and arranged images is shown at Table 2. After arrangement, the number of each class was ranged from 39 to 1223.

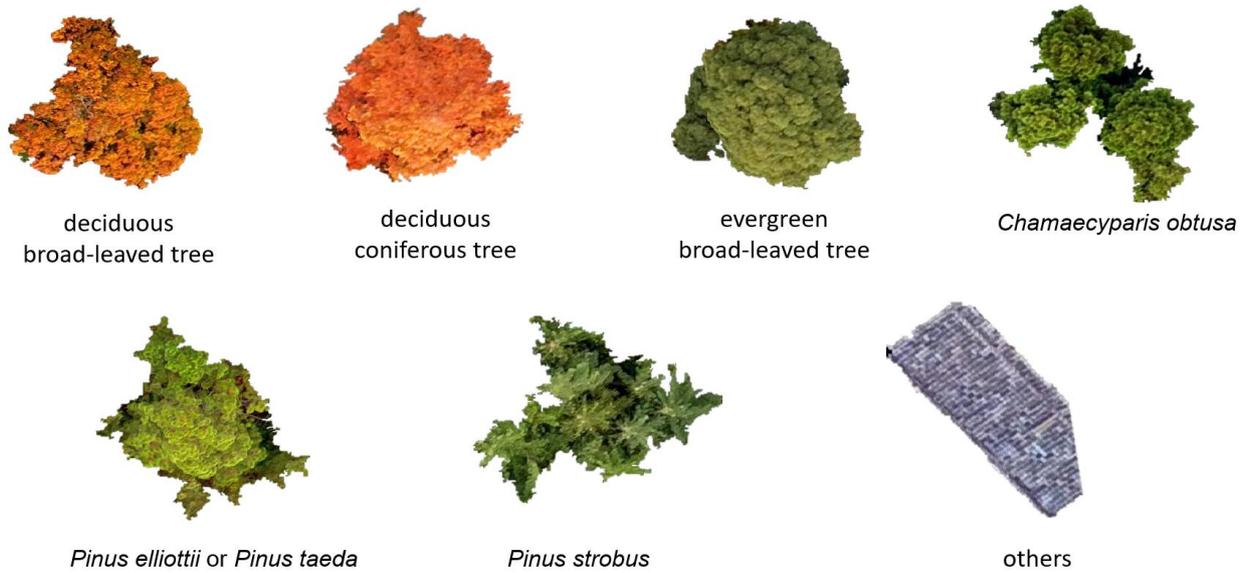

**Figure 3.** Extracted each class sample

**Table 2.** Number of each class images

|   | extracted images | arranged images | Training and validation images | increased training and validation images | test images |
|---|---|---|---|---|---|
| 1 | 202 | 195 | 147 | 3087 | 48 |
| 2 | 74 | 53 | 40 | 840 | 13 |
| 3 | 245 | 100 | 75 | 1575 | 25 |
| 4 | 957 | 348 | 261 | 5481 | 87 |
| 5 | 260 | 206 | 155 | 3255 | 51 |
| 6 | 104 | 39 | 30 | 630 | 9 |
| 7 | 1419 | 1223 | 918 | 918 | 305 |

class1: deciduous broad-leaved tree, 2: deciduous coniferous tree 3: evergreen broad-leaved tree, 4: *Chamaecyparis obtuse*, 5: Pinus, 6: *Pinus strobus*, 7: others

## 2.4. Deep Learning

We separated 50% of arranged each class images to training dataset, 25% to validation dataset and 25% to test dataset randomly. To make a model for object identification, we used a publicly available package nVIDIA DIGITs 5.0 (Heinrich 2016) as a deep learning framework and GoogLeNet (Szegedy et al. 2015) as a transfer learning model. To fit the image sizes to 256*256 pixels, we filled the surrounding pixels with certain character. Other learning settings were basically default to the DIGITs Image Classification Model (Table 3).

Table 3. The setting of the DIGITS Image Classification Model

| Setting | Selected option |
|---|---|
| Training epochs | 30 |
| Snapshot interval | 1 |
| Validation interval | 1 |
| Random seed | None |
| Batch size | Network defaults |
| Solver type | SGD |
| Base learning rate | 0.01 |
| Policy | Step down |
| Step size | 33 |
| Gamma | 0.1 |
| Network | GoogLeNet |

## 3. Results

First, we made model 1 which is trained by the arranged images. The performance of Model 1 is relatively favorable. Then, we increased the number of training and validation images and made Model 2. Model 2 showed a significant improvement compared to Model 1.

The performance of the model 1 is shown at Table 4. An overall accuracy was 83.1%. However, model 1 was not able to identify "deciduous coniferous tree", "evergreen broad-leaved tree", and "*Pinus strobus*". The "deciduous coniferous tree" was mostly misclassified into "deciduous broad-leaved tree".

Table 4. Confusion matrix of model 1. Vertical axis is ground truth and horizontal axis is model prediction.

|   | 1 | 2 | 3 | 4 | 5 | 6 | 7 | Per-class accuracy |
|---|---|---|---|---|---|---|---|---|
| 1 | 41 | 0 | 0 | 0 | 5 | 0 | 2 | 85.42% |
| 2 | 11 | 0 | 0 | 0 | 0 | 0 | 2 | 0.0% |
| 3 | 0 | 0 | 5 | 4 | 0 | 0 | 16 | 20.0% |
| 4 | 0 | 0 | 5 | 73 | 3 | 0 | 6 | 83.91% |
| 5 | 6 | 0 | 0 | 8 | 30 | 0 | 7 | 58.82% |
| 6 | 0 | 0 | 2 | 7 | 0 | 0 | 0 | 0.0% |
| 7 | 4 | 0 | 1 | 1 | 1 | 0 | 298 | 97.7% |
| overall | | | | | | | | 83.1% |

class1: deciduous broad-leaved tree, 2: deciduous coniferous tree 3: evergreen broad-leaved tree, 4: *Chamaecyparis obtuse*, 5: Pinus, 6: *Pinus strobus*, 7: others

The cause of low accuracies of these classes was thought to be the small number of training images, so we increased training and validation images. We made 21 images from 1 image using random rotation, width and height shift, shear strain, zoom and flip. We applied this method to each class, except for "others" class because this class had already many images, and made model 2 which was trained by these large number of images.

Model 2 has improved the performance significantly (Table 5). The overall accuracy was improved to 89.0% and most class achieved 85% ~ 95% accuracy. Compared to model 1, "deciduous coniferous tree", "evergreen broad-leaved tree" and "*Pinus strobus*" class was clearly classified at around 90% accuracy.

Table 5. Confusion matrix of model 2 trained by increased images.

|   | 1 | 2 | 3 | 4 | 5 | 6 | 7 | Per-class accuracy |
|---|---|---|---|---|---|---|---|---|
| **1** | 46 | 2 | 0 | 0 | 0 | 0 | 0 | 95.83% |
| **2** | 2 | 11 | 0 | 0 | 0 | 0 | 0 | 84.62% |
| **3** | 0 | 0 | 17 | 2 | 1 | 0 | 5 | 68.0% |
| **4** | 0 | 0 | 3 | 80 | 4 | 0 | 0 | 91.95% |
| **5** | 0 | 0 | 1 | 1 | 48 | 0 | 1 | 94.12% |
| **6** | 0 | 0 | 0 | 0 | 1 | 8 | 0 | 88.89% |
| **7** | 8 | 1 | 20 | 1 | 5 | 1 | 269 | 88.2% |
| **overall** | | | | | | | | 89.0% |

class1: deciduous broad-leaved tree, 2: deciduous coniferous tree 3: evergreen broad-leaved tree, 4: *Chamaecyparis obtuse*, 5: Pinus, 6: *Pinus strobus*, 7: others

## 4. Discussion

The objective of this research was to test whether our machine vision system can classify individual trees into several tree types (deciduous and coniferous) and identify a few specific tree species. Focusing on tree type, model 2 classified 4 tree types at 92.7% accuracy. At species level, model 2 was able to classify *Chamaecyparis obtuse*, *Pinus strobus* from *Pinus elliottii* and *Pinus taeda* at high accuracy. These results mean our system is able to classify tree types and have a potential to classify tree species.

This performance is notable because we only used easily available digital RGB images and publicly available package for deep learning. In contrast, most of previous studies used expensive hardware such as multispectral imagers to improve performance. In the matter of spatial scale, our method using a UAV can be limited more than previous method using airborne. But low-cost and easy-to-use feature of UAVs can enable us to periodic monitoring. Our machine vision system will be a cost-effective and usable tool for forest remote sensing.

While we achieved high accuracy classification, some misclassification also exists. The main error is classification between "evergreen broad-leaved tree" and "others" class. This may be caused by the fact that "others" class included some understory vegetation. Understory vegetation mainly consisted of evergreen broad-leaved tree. How we separate each class is one of the key parameters which affect the results.

In our method, there are two reasons for the fine result. First one is that we conducted object-based classification. Previous study shows object-based classification get higher accuracy than pixel-based classification (Tarabalka et al. 2010). In this matter, our method could not segment every each tree crown apart perfectly, so improving the segmentation method will lead to higher accuracy classification, and it enable us to count the number of each tree species.

    The other reason for the fine result is that we picked up training and test images from the same area and the same time. Tree shapes are thought to be different in different environments, and leaf colors and illuminations are different at season and weather. Making a good use of tree shapes (or DEM) and seasonality of leaf colors will improve classification accuracy, but generally these properties may have a bad influence for simple machine learning. Considering practicability, versatile model which is trained images of various site and time is desired in the further study.

**Literature Cited**

Baatz, M. and Schape, A. (2000). Multiresolution Segmentation: An Optimization Approach for High Quality Multi-Scale Image Segmentation. Angewandte Geographische Informationsverarbeitung, 12, 12-23.

Fassnacht, F. E., Latifi, H., Stereńczak, K., Modzelewska, A., Lefsky, M., Waser, L. T., ... & Ghosh, A. (2016). Review of studies on tree species classification from remotely sensed data. *Remote Sensing of Environment*, *186*, 64-87.

Heinrich, G. (2016). Image Segmentation Using DIGITS 5.
URL https://devblogs.nvidia.com/image-segmentation-using-digits-5/

Heinzel, J., & Koch, B. (2012). Investigating multiple data sources for tree species classification in temperate forest and use for single tree delineation. *International Journal of Applied Earth Observation and Geoinformation*, *18*, 101-110.

Ise, T., Minagawa, M., and Onishi, M. (2018). Identifying 3 moss species by deep learning, using the" chopped picture" method. *Open Journal of Ecology 08(03)* doi: 10.4236/oje.2018.83011

Krizhevsky, A., Sutskever, I., & Hinton, G. E. (2012). Imagenet classification with deep convolutional neural networks. In *Advances in neural information processing systems* (pp. 1097-1105).

Machala, M., & Zejdová, L. (2014). Forest mapping through object-based image analysis of multispectral and LiDAR aerial data. *European Journal of Remote Sensing*, *47*(1), 117-131.

Paneque-Gálvez, J., McCall, M. K., Napoletano, B. M., Wich, S. A., & Koh, L. P. (2014). Small drones for community-based forest monitoring: An assessment of their feasibility and potential in tropical areas. *Forests*, *5*(6), 1481-1507.

Peerbhay, K. Y., Mutanga, O., & Ismail, R. (2013). Commercial tree species discrimination using airborne AISA Eagle hyperspectral imagery and partial least squares discriminant analysis (PLS-DA) in KwaZulu–Natal, South Africa. *ISPRS Journal of Photogrammetry and Remote Sensing*, *79*, 19-28.

Szegedy, C., Liu, W., Jia, Y., Sermanet, P., Reed, S., & Anguelov, D. & Rabinovich, A. (2015). Going deeper with convolutions. In *Proceedings of the IEEE conference on computer vision and pattern recognition* (pp. 1-9).

Tang, L., & Shao, G. (2015). Drone remote sensing for forestry research and practices. *Journal of Forestry Research*, *26*(4), 791-797.

Tarabalka, Y., Fauvel, M., Chanussot, J., & Benediktsson, J. A. (2010). SVM-and MRF-based method for accurate classification of hyperspectral images. *IEEE Geoscience and Remote Sensing Letters*, *7*(4), 736-740.